\ifdefined\XeTeXversion\else\pdfoutput=1\fi
\documentclass{article}

\usepackage{iclr2027_conference,times}
\iclrfinalcopy %
\usepackage[T1]{fontenc}
\usepackage[utf8]{inputenc}
\usepackage{microtype}
\usepackage{amsmath,amssymb}
\usepackage{graphicx}
\usepackage{float}
\usepackage{booktabs}
\usepackage{multirow}
\usepackage{xcolor}
\usepackage{tabularx}
\usepackage{colortbl}
\definecolor{paperblue}{HTML}{0072B2}
\newcommand{\gaterow}{\rowcolor{paperblue!10}}
\usepackage[colorlinks=true,linkcolor=paperblue,citecolor=paperblue,urlcolor=paperblue]{hyperref}

\graphicspath{{figures/}}

\title{Local or Escalate?\\
Post-Training Expert Routing for\\
Real-Time Duplex Speech Models}

\author{%
  Ruoyu He$^{1,*,\dagger}$, Jisen Li$^{2,*}$, Anthony Ji$^{3}$, Changyi Yang$^{4}$,\\
  \textbf{Zetian Li$^{5}$, Runduo Ma$^{6}$, Zhen Zhang$^{7}$, Zhongzhu Zhou$^{8}$}\\[0.5ex]
  $^{1}$University of California, Los Angeles\quad
  $^{2}$University of Illinois at Urbana-Champaign\\
  $^{3}$University of British Columbia\quad
  $^{4}$Carnegie Mellon University\quad
  $^{5}$Stanford University\\
  $^{6}$University of Wisconsin--Madison\quad
  $^{7}$University of California, Santa Barbara\\
  $^{8}$University of Sydney
}

\begin{document}
\maketitle
\lhead{}
\renewcommand{\headrulewidth}{0pt}
{\renewcommand{\thefootnote}{*}\footnotetext{Equal contribution. \textsuperscript{\ensuremath{\dagger}}Corresponding author: Ruoyu He (\href{mailto:rhe9527@gmail.com}{\texttt{rhe9527@gmail.com}}).}}

\begin{abstract}
A full-duplex model should answer easy questions immediately, but
hard questions often need stronger reasoning, which may require waiting. The challenge
is to choose between these paths at the moment the model begins to speak.
Recent systems learn related mode decisions through special-token generation
inside the dialogue model, coupling routing to turn-taking and temporally
annotated full-duplex training. We study a post-training alternative. We freeze
MiniCPM-o~4.5, using only its speech input and output, and fit a logistic
failure readout on hidden states already
available at its native listen-to-speak transition. The resulting gate requires
no additional model generation and exposes thresholds that can be retuned
without changing the duplex policy. A probe fitted on text-input states also
reads the states of spoken questions, including unseen human recordings.
On the internal benchmark, selective handoff improves answer accuracy by 27.1
points. Compared with always using the expert, it reduces mean server time to
first answer audio by 31.5 percent. It also outperforms matched random routing
across external speech question-answering pools, as well as a zero-shot
GPT-5.4 mini router that rates the transcribed question. The training recipe also transfers to a second duplex model family. These results show that an existing duplex model can be
equipped to answer easy turns immediately and wait for stronger reasoning or
fresher knowledge on the turns that need it, without further duplex training.
\end{abstract}

\section{Introduction}
\label{sec:intro}
Real-time conversation rewards two behaviors that pull in opposite directions.
For a greeting or an easy factual question, a speech agent should answer as
soon as it understands the turn. For a difficult knowledge or reasoning
question, speaking immediately can lock the interaction into a fluent but
incorrect answer, so taking time to reason or ask a stronger model is preferable.
The choice is therefore whether to answer now or wait for stronger
reasoning or fresher knowledge.

Native duplex models already learn a fine-grained listen/speak policy
\citep{defossez2024moshi,openbmb2026minicpmo}. Recent systems extend this idea
with tokenized control. BayLing-Duplex uses four dialogue-state tokens for
turn-taking and interruption \citep{bayling2026duplex}. DuplexOmni uses a
\texttt{[THINK]} token to request an asynchronous reasoning layer
\citep{huang2026duplexomni}, and Listen-Write-Speak encodes concurrent
listening, writing, and speaking with a token schema
\citep{zhang2026liberating}. These control decisions are learned alongside
speech content, timing, and interruption. DuplexOmni and Listen-Write-Speak
construct large temporally annotated corpora to train this invocation
behavior within the interaction model.

We ask whether failure risk can be read from the model at the moment it
takes the floor. Readiness to speak does not ensure the knowledge needed
to answer correctly, while routinely invoking an expert makes easy turns
wait. A useful failure-risk estimate should be available at the native speak
boundary, preserve the learned duplex policy, and expose an adjustable
accuracy--latency operating point.

Hidden-state probes can predict correctness in
text models \citep{azaria2023internal,kossen2024semantic,lugoloobi2026failures},
but duplex speech introduces three shifts: audio input, representations
adapted for interaction, and a decision deadline at answer onset. We test
whether useful failure information survives these shifts in a frozen model
and supports a serving policy before a complete answer can be inspected.

\begin{figure}[t]
  \centering
  \includegraphics[width=\linewidth]{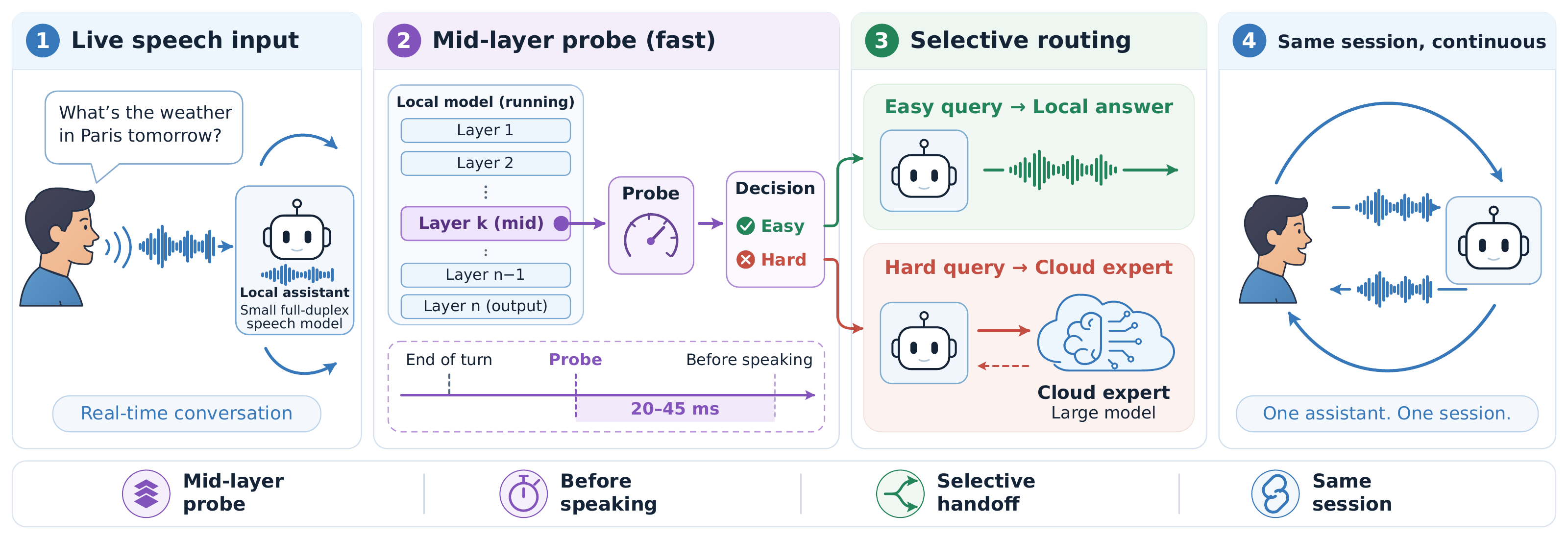}
  \caption{\textbf{Post-training routing at the native speak boundary.}
  \textbf{(1)} Speech streams in 1\,s chunks to a frozen full-duplex local
  model. \textbf{(2)} When the model commits to speak, an L22 linear readout
  scores local failure risk in tens of milliseconds. \textbf{(3)} Low-risk turns continue
  immediately, while high-risk turns are sent to a stronger expert and the
  talker holds the floor. \textbf{(4)} The expert answer is voiced by the local
  talker. On all 240 internal queries, the aggressive policy raises accuracy
  from 42.1\% to 69.2\% at 50.0\% escalation.}
  \label{fig:hero}
\end{figure}

Our system freezes MiniCPM-o~4.5 \citep{openbmb2026minicpmo}, an omni-modal
full-duplex model used here for speech only, and fits a regularized logistic
readout on a few thousand cached answer-onset states. This separates
\emph{interaction control} (when the model takes the floor) from
\emph{competence control} (whether to trust its answer). The gate adds no
language-model forward pass or duplex training, and its threshold adjusts
expert use without changing the checkpoint. Retained turns continue locally;
on escalated turns, the local talker voices the expert's answer.

Our experiments connect representation analysis to serving behavior. Probes
fitted on text states read unseen human speech, while layer controls reveal
that the late-layer decline depends on input modality
(\S\ref{sec:findings}). At the system level, selective handoff improves
internal accuracy by 27.1 points over local answering and reduces mean server
waiting by 31.5\% relative to always escalating. Gains over matched random
routing extend across four external speech-QA pools. In a blinded live study,
13 participants preferred the gated system in most paired conversations,
especially on real-time-information tasks (\S\ref{sec:human-eval}). Together,
these results support allocating expert access at the native speak boundary
while retaining the frozen model's interaction policy.

\section{Related Work}
\label{sec:related}
\paragraph{Learned control in duplex speech models.}
Native spoken-dialogue models coordinate concurrent listening and speaking
\citep{defossez2024moshi,openbmb2026minicpmo}, using learned states,
synchronization mechanisms, or lightweight adapters
\citep{wang2024fsm,zhang2024duplex,veluri2024syncllm,wang2024freezeomni}.
BayLing-Duplex interleaves user audio, assistant text, and assistant audio. Four
special tokens reduce silence, reply onset, text completion, and speech
completion to next-token prediction \citep{bayling2026duplex}. MiniCPM-o~4.5
similarly predicts a binary Listen/Speak token before content. These signals
control the floor, not whether the forthcoming answer is correct.

\paragraph{Reasoning while interacting.}
Chronological Thinking reasons incrementally while listening, then stops
thinking when speaking begins \citep{chrono2026thinking}.
Listen-Write-Speak uses a token schema for an
always-on visible writing channel alongside listening and speaking
\citep{zhang2026liberating}. DuplexOmni already separates interaction from
thinking: its interaction model emits \texttt{[THINK]} to invoke a pluggable
background reasoner and learns delayed-result integration from
Writer--Director data \citep{huang2026duplexomni}. MoshiRAG fine-tunes Moshi
to emit a learned \texttt{<ret>} token that triggers retrieval, exploiting
the delay before key information is spoken to retrieve asynchronously
\citep{chien2026moshirag}. In both systems the
decision to deliberate is a token that the interaction model is trained to
emit. Our gate instead reads failure risk from the native answer-onset states
of a frozen duplex checkpoint, before a completed local answer, and routes the
answer to an external expert, leaving the checkpoint unchanged. To our
knowledge, no prior duplex system derives this handoff decision from the
frozen model's own hidden states and measures its effect in live deployment.

\paragraph{Interactive serving infrastructure.}
Complementary to model-level interaction and reasoning control, SGLang-Omni
provides a unified runtime for native full-duplex, turn-based, and cascaded
interactive systems. Composable graphs connect models, tools, and interaction
control, while persistent sessions coordinate their state and lifecycles
\citep{sglangomniinteractive2026}. This infrastructure supports continuous
interaction and asynchronous execution; our work addresses the complementary
competence decision of when a frozen duplex model should delegate an answer
to a stronger expert, using its native answer-onset states.

\paragraph{Competence readout and selective routing.}
Correctness can be estimated from verbal self-evaluation
\citep{kadavath2022language,lin2022teaching}, hidden representations
\citep{azaria2023internal,burns2023discovering,marks2024geometry}, or
uncertainty across sampled answers \citep{kuhn2023semantic,farquhar2024detecting}.
Internal representations can encode correct answers absent from model outputs
\citep{orgad2025llms}, and trained hidden-state probes can outperform other
factual-confidence estimators \citep{mahaut2024factual}.
Semantic Entropy Probes amortize
sample-based uncertainty into a hidden-state read \citep{kossen2024semantic}.
Text routers allocate queries among models of different cost
\citep{chen2023frugalgpt,ding2024hybrid,ong2024routellm}, including from
prefill activations \citep{varshney2026llmrouter,lugoloobi2026failures}.
A linear probe is therefore not our novelty. Our contribution is to make such a
readout work at a native audio listen-to-speak boundary, study its survival
across speech and duplex transformations, and measure the resulting
accuracy--latency policy in a live speech pipeline with end-to-end expert
handoff.

\section{Post-Training Local-or-Escalate Routing}
\label{sec:signal}

\subsection{The local--expert loop}
\label{sec:system}
Selective model routing balances response quality against the cost
of a stronger model \citep{ong2024routellm}. We place that decision at
a local full-duplex speech model's transition from listening to
speaking (Figure~\ref{fig:hero}). The local model retains its native
decision about when to respond, and the gate decides whether the response
should begin immediately or wait for expert help. On the expert path,
an audio uplink transcribes the question, the expert returns a text
answer, and the local talker voices the prepared relay. Until the relay
starts, the talker holds the floor with one fixed line:
``Hmm, let me double-check that---one moment.'' The line is synthesized
once in the talker's voice, played when the gate fires, and written into
the duplex context as the model's own speech. After the line, the model listens in silence until
the relay starts, with no repeated filler. We use predicted
local-answer failure as the routing score and vary a threshold to expose
the accuracy--latency frontier across different expert-call budgets.

\subsection{A probe for the handoff decision}
\label{sec:probe}
Linear \emph{probes} test what information a classifier can read from
frozen representations \citep{alain2016understanding}. Hidden-state
classifiers have estimated statement truthfulness
\citep{azaria2023internal} and pre-generation task success
\citep{lugoloobi2026failures}. We use this approach to ask whether a
speech model's existing states contain a directly readable failure
signal. Reusing these states avoids a separate self-evaluation
generation and leaves the speech model unchanged.

Offline, we collect local answers to calibration questions and use
a judge to assign correctness labels. Each label is paired with
hidden-state features captured at the chosen read point. We fit a
logistic probe to predict failure from these features while keeping
the speech model frozen. For features $\phi(x)$ from input context
$x$, the fitted weights $w,b$ give the risk score
\begin{equation}
 s(x)=\sigma\!\left(w^\top\phi(x)+b\right).
 \label{eq:gate}
\end{equation}
At inference, the probe reads the current features without a
reference answer or an online judge. Scores above a
language-specific threshold $\tau_{\ell}$ favor escalation.
The thresholds select nominal expert-call budgets from calibration scores
and stay fixed during evaluation (\S\ref{sec:models}). Realized call
rates can change when the query distribution shifts. An auxiliary dialogue-act head
restricts escalation to information requests, filtering turns such
as acknowledgments and stop commands (Appendix~\ref{app:actgate}).

\paragraph{Why separate routing from generation?}
Only $w$ and $b$ are fitted, and the regularized logistic objective is convex and
uses cached features and answer labels. It therefore cannot alter the local
model's answer distribution, listen/speak behavior, or speech quality. The
operating threshold is a serving parameter rather than a generated token, so it
can be recalibrated for a new expert budget without another duplex fine-tune.
This is the precise training-stability advantage of our design: it removes
competence arbitration from the joint autoregressive objective rather than
claiming that special-token control cannot be trained successfully.

\begin{figure}[!t]
  \centering
  \includegraphics[width=\linewidth]{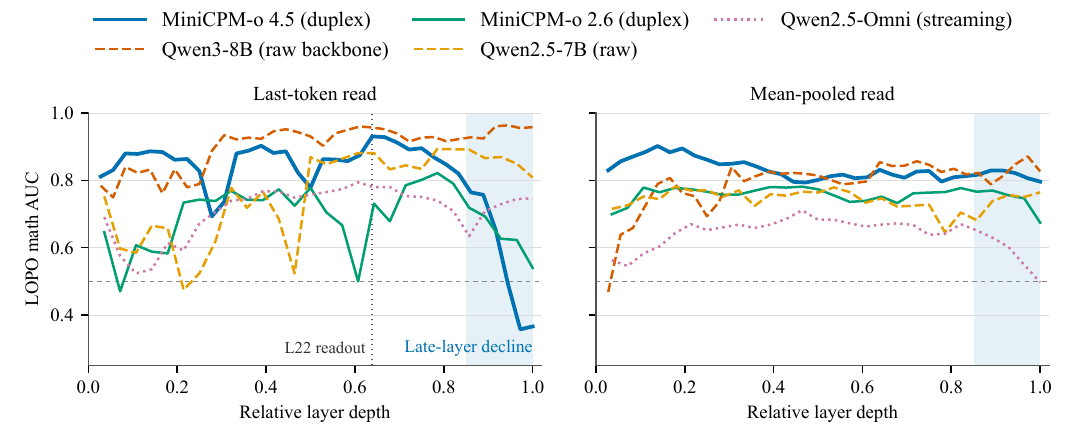}
  \caption{\textbf{Text-input transfer depends on layer and position.}
  Held-out-pool (LOPO) math AUC of per-layer failure probes, read at the
  last token (left) or mean-pooled (right). The dotted line marks the
  deployed L22 readout, and \S\ref{sec:finding-cliff} discusses the
  late-layer decline.}
  \label{fig:layersweep}
\end{figure}

\subsection{Where should the probe read?}
\label{sec:readlocation}
Adapting the probe to duplex speech requires choosing its layer,
position, and read time. The final layer is optimized for the output
head, whereas an earlier layer can retain more transferable task
information. Figure~\ref{fig:layersweep} motivates L22 through
text-input leave-one-pool-out (LOPO) evaluation: on held-out math,
its last-token probe reaches 0.931 AUC, versus 0.366 at the final
layer. The contrast is regime-specific rather than a universal
middle-layer advantage: it disappears when both calibration and
inference use audio input in the matched native setting
(\S\ref{sec:finding-cliff} and Appendix~\ref{app:signal}).

The native feature vector concatenates the latest L22 tail state,
the mean of the tail states, and the mean input state, allowing the
probe to combine the current boundary state with pooled context.

We read these features when the generation call first returns a
listen-to-speak transition. This makes the gate available at the
local model's response boundary without waiting for a completed
answer. The call can already contain the first generated text unit,
so we call the decision point \emph{answer onset}. Further answer
chunks may expose additional failure information, at the cost of a
later handoff (Appendix~\ref{app:readtime}). We fit the native probe
and thresholds on features and answer labels from this serving regime
to match its audio input and generation behavior. These choices
raise two questions that Section~\ref{sec:findings} explores separately
from the native calibration: why only text input loses the final-layer
signal, and whether a probe fitted on text states can read spoken
questions.

\section{Experiments}
\label{sec:setup}
\label{sec:results}
\subsection{Experimental setup}
\label{sec:models}
\label{sec:dataset}
\label{sec:signals}
\label{sec:evalproto}
\label{sec:liveproto}
\label{sec:inject}
We evaluate frozen MiniCPM-o~4.5 (9B) as a speech-only duplex model (audio
in, speech out, no image or video input), with \texttt{gpt-5.5}, which can
call web search, as the escalation expert, on the full 240-query internal test set and four
external English speech-QA pools: Speech TriviaQA, Speech Web Questions,
and Speech Llama Questions from OpenAudioBench \citep{li2025baichuanaudio},
and SD-QA from VoiceBench \citep{faisal2021sdqa,chen2024voicebench}.
The deployed probe is fitted on 5,228 calibration rows. Per-language thresholds use
core-row out-of-fold scores and stay fixed across pools.
Calibration details and model controls are in
Appendix~\ref{app:current-protocol}.

Each query has one sampled native session per arm: local-only,
conservative, balanced, aggressive, and always-escalate.
We report \emph{answer-content accuracy}, scored from generated local
text or intended relay text. The random reference is the analytical
mixture $A_{\rm mix}(r)=(1-r)A_{\rm local}+rA_{\rm always}$ at the
gate's realized call rate $r$. Both model families use all 240
Internal query IDs: chat (60), factual QA (40), knowledge (60), math (60),
and SimpleQA traps (20). Every query has unit weight in accuracy and
call-rate calculations. Table~\ref{tab:live} gives the per-tier results
and uncertainty intervals on this full test set.
External averages weight the four QA pools equally.
\paragraph{Unified sessions and timing endpoint.}
Accuracy and latency are measured on the same 5,950 recorded sessions
across the five pools and five policies (the full query audio
is transcribed for the expert, never before its last sample's
scheduled arrival), and TTFA comes directly from session timestamps.
Appendix~\ref{app:metric-sources} gives the input and recording
details for each metric. Server TTFA measures the wait
from scheduled input end until the first answer PCM block is ready,
with early responses assigned zero wait. Means include every completed
session with a recorded answer timestamp, including long waits; failures
have no TTFA value and count as incorrect for accuracy.
Both routing methods start ASR at the known end of the replayed input.
This measures server waiting under controlled replay, without measuring
live endpoint detection or client playback. The holding line
(\S\ref{sec:system}) is excluded.
Appendix~\ref{app:setup-details} provides the model, probe-fitting,
and answer-judging configurations used in these experiments.

\begin{table}[t]
\centering\small
\setlength{\tabcolsep}{3pt}
\renewcommand{\arraystretch}{1.12}
\caption{\textbf{Answer-content accuracy (\%), time, and expert use.}
Both models use 240 Internal IDs and four external QA pools. The avg column is the
external mean and $\Delta$ its gain over local. Random rows match each
aggressive call rate.
Expert USD/query is estimated by replaying escalated uplink transcripts
through the identical expert configuration at official rates
(Appendix~\ref{app:expertcost}).}
\label{tab:transfer}
\begin{tabularx}{\linewidth}{@{}l*{7}{>{\raggedleft\arraybackslash}X}@{}}
\toprule
& Internal & TriviaQA & WebQ & Llama Q. & SD-QA & avg & $\Delta$ \\
\midrule
\multicolumn{8}{l}{\emph{MiniCPM-o~4.5}} \\
always-local & 42.1 & 54.0 & 54.8 & 79.6 & 49.0 & 59.4 & --- \\
\gaterow\;+ gate, conservative & 48.8 & 57.6 & 58.0 & 78.8 & 52.0 & 61.6 & +2.3 \\
\gaterow\;+ gate, balanced & 60.0 & 68.8 & 66.0 & 82.0 & 67.0 & 71.0 & +11.6 \\
\gaterow\;\textbf{+ gate, aggressive} & \textbf{69.2} & \textbf{87.2} & \textbf{75.2} & \textbf{87.2} & \textbf{83.0} & \textbf{83.2} & \textbf{+23.8} \\
\;matched random, aggressive & 60.8 & 77.2 & 69.6 & 82.7 & 74.6 & 76.0 & +16.7 \\
always-escalate & 79.6 & 96.0 & 80.8 & 94.0 & 90.0 & 90.2 & +30.9 \\
\midrule
\multicolumn{8}{l}{\emph{VoiceChat-11B}} \\
always-local & 19.6 & 35.8 & 37.6 & 70.8 & 31.0 & 43.8 & --- \\
\gaterow\;+ gate, conservative & 31.7 & 48.4 & 48.8 & 80.7 & 42.0 & 55.0 & +11.2 \\
\gaterow\;+ gate, balanced & 44.6 & 61.0 & 59.2 & 85.8 & 54.0 & 65.0 & +21.2 \\
\gaterow\;\textbf{+ gate, aggressive} & \textbf{60.4} & \textbf{76.8} & \textbf{69.6} & \textbf{89.7} & \textbf{66.5} & \textbf{75.7} & \textbf{+31.9} \\
\;matched random, aggressive & 53.5 & 65.7 & 60.2 & 82.0 & 60.0 & 67.0 & +23.2 \\
always-escalate & 91.7 & 95.5 & 82.8 & 93.1 & 89.0 & 90.1 & +46.3 \\
\bottomrule
\end{tabularx}
\vspace{4pt}
\begin{tabular*}{\linewidth}{@{\extracolsep{\fill}}lrrrrr@{}}
\toprule
\multicolumn{6}{l}{\emph{MiniCPM cost summary, Internal set (all metrics lower is better)}} \\
& local & conservative & balanced & aggressive & always \\
\midrule
Mean TTFA (s) & 1.08 & 2.15 & 4.67 & 5.95 & 8.68 \\
GPT expert USD/query & 0 & 0.003 & 0.009 & 0.012 & 0.018 \\
\bottomrule
\end{tabular*}
\end{table}

\begin{figure}[!t]
  \centering
  \includegraphics[width=0.9\linewidth]{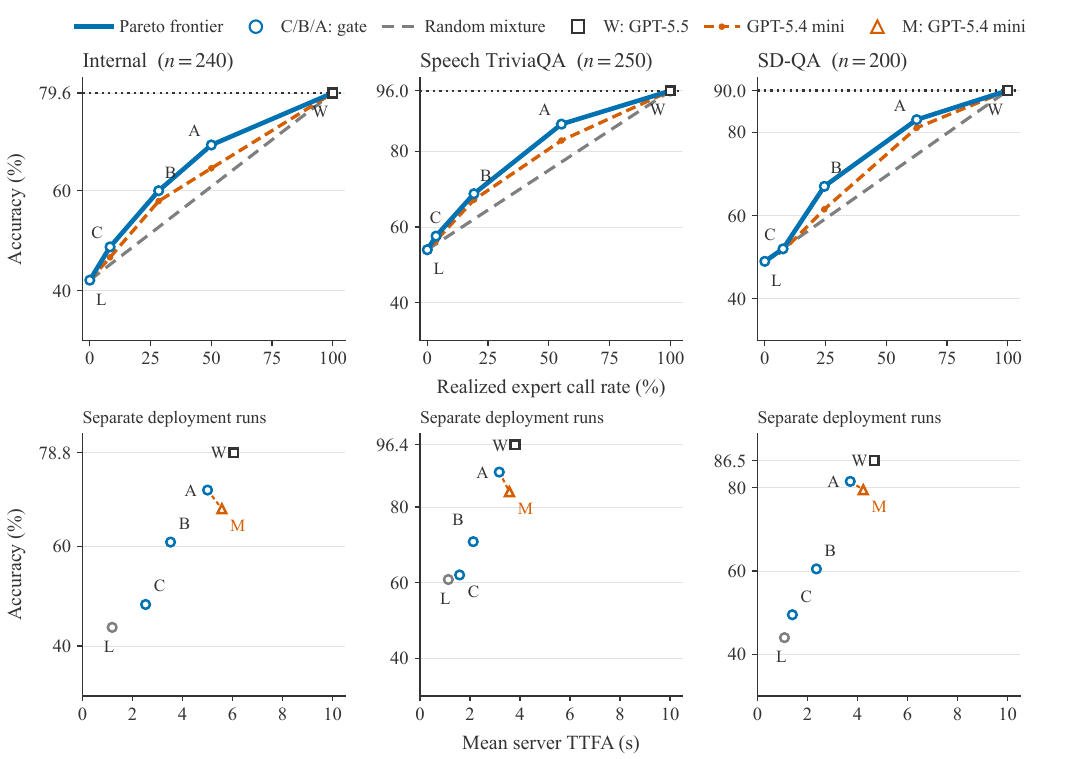}
  \caption{\textbf{Accuracy--cost trade-offs across routing policies.}
  Top row: accuracy versus expert call rate in the main run.
  Bottom row: accuracy versus mean server TTFA in separate deployment
  runs on the same query sets; two anomalous samples are excluded.
  L/C/B/A/W denote local,
  conservative, balanced, aggressive, and always-expert (GPT-5.5)
  policies. Dashed orange: the zero-shot GPT-5.4 mini router applied to
  the same sessions at the gate's C, B and A call rates. M: the same
  router run live as the escalation decision at the aggressive budget.
  Each row uses its own judged answers, so W accuracy can differ between
  rows. Accuracy includes failed attempts; TTFA requires an answer
  timestamp. Run sources, denominators, and a 60\,s exclusion sensitivity
  analysis are in Appendix~\ref{app:llmdeploy}.}
  \label{fig:dualview}
\end{figure}

\begin{figure}[t]
\centering
\includegraphics[width=0.9\linewidth]{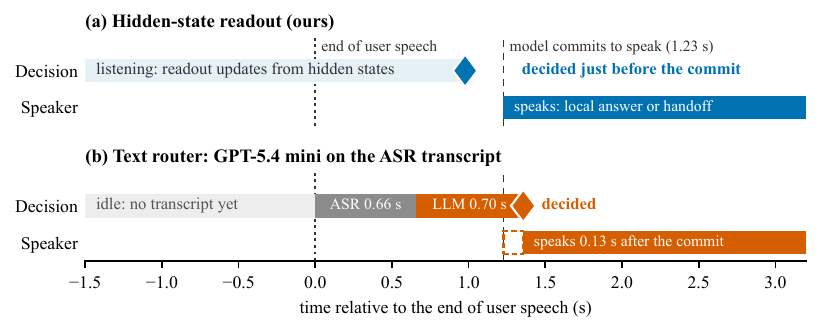}
\caption{\textbf{When the escalation decision is made} (medians, aggressive
run). (a) The readout decides when the model commits to speak. (b) A text
router needs the transcript and an LLM call, so it decides after the commit
even with the call started as soon as the transcript is ready. Live, it adds
0.64\,s on turns both policies answer locally (Appendix~\ref{app:llmdeploy}).}
\label{fig:llmtimeline}
\end{figure}

\subsection{Internal accuracy and timing cost}
\label{sec:live}
\label{sec:liveloop}
\label{sec:livecurve}
On the full 240-query internal set (upper block of
Table~\ref{tab:transfer}), the aggressive gate escalates half of the
queries and raises accuracy from 42.1\% for local answering to
\textbf{69.2\%}. This is significantly above matched random routing
(+8.3 points) and recovers 72.2\% of the gap to always escalating. The
conservative and balanced tiers fall in between (Table~\ref{tab:live}).
The gain is concentrated on factual QA and questions with misleading
premises, where the gate can tell that answering at once is risky
(Appendix~\ref{app:internal-strata}).

Waiting time is reported in Table~\ref{tab:latency}.
Compared with always escalating, the aggressive gate keeps 86.9\% of the
accuracy (69.2\% vs.\ 79.6\%) and cuts mean server waiting time by
31.5\%, from 8.68\,s to 5.95\,s, without retraining the duplex model.
Turns that stay local wait 1.23\,s on average, so fast interaction is
preserved on them. Escalated turns wait 10.58\,s, mostly for the
expert's own response rather than for serving overhead. Medians, tails,
and early responses are in Table~\ref{tab:latency}, and
Appendix~\ref{app:ttfa-decomp} splits the wait into expert time and
system overhead.

\subsection{External transfer and its limits}
\label{sec:transfer}
\label{sec:transfer-neg}
The native gate's weights, feature definition, layer, and per-language
thresholds stay fixed across the four external English QA pools. On
these pools the aggressive policy raises macro accuracy from
\textbf{59.4\% to 83.2\%}, 7.1 points above matched random routing.
A replay with each query's local and expert answers fixed gives a
similar gain (Appendix~\ref{app:pairedci}).
Realized call rates adapt to each pool, from 21.2\% on Llama Questions
to 62.5\% on SD-QA, and the largest gains appear on TriviaQA and SD-QA.

A TF--IDF router fitted on the same rows transfers much worse than the
readout (Appendix~\ref{app:textrouter}). Next-token entropy at the read
point is near chance, and non-linear heads do not beat the linear readout
(Appendix~\ref{app:confbaselines}). A fast
LLM, \texttt{gpt-5.4-mini}, used as is, can instead rate the live ASR
transcript for whether a small model would get the question wrong. As
the escalation decision, this zero-shot router is less accurate and
adds waiting on locally answered turns. In a replay that ranks the same sessions of the main run
with every answer fixed, it ranks local failures worse than the readout,
and the gate is more accurate at the aggressive call rate
(+2.6 points, dashed in Figure~\ref{fig:dualview}). Run live as the
escalation decision on three pools, the router has 77.3\% macro accuracy
against the gate's 80.8\%, and on the turns that both
answer locally it adds 0.64\,s before the answer (M in
Figure~\ref{fig:dualview}, timing in Figure~\ref{fig:llmtimeline}).
The pooled mean over all paired turns is essentially equal for the two
methods; the latency advantage is specific to the local-answer path.
It is also weaker on the calibration mix and
on Chinese questions. Appendix~\ref{app:llmdeploy} gives the results by
pool and the other latency summaries.

Specific-but-wrong answers make up 70\% of failures, and the readout
ranks them well (AUC 0.81). Execution slips in multi-step reasoning are
near chance (0.47). Reading two answer chunks later raises external mean
AUC from 0.755 to 0.791, at the cost of a later handoff
(Appendix~\ref{app:readtime}).

The lower block of Table~\ref{tab:transfer} repeats the recipe on
NemotronLabs-VoiceChat-11B, refitted rather than reusing the weights,
with its layer set chosen with the Internal split in view. In cached
replay at a 50\% budget, external macro accuracy rises from 43.8\% to
75.7\%, against 67.0\% for matched random routing
(Appendix~\ref{app:nvda}).

\section{Exploratory Analysis}
\label{sec:findings}

\begin{figure}[t]
  \centering
  \includegraphics[width=\linewidth]{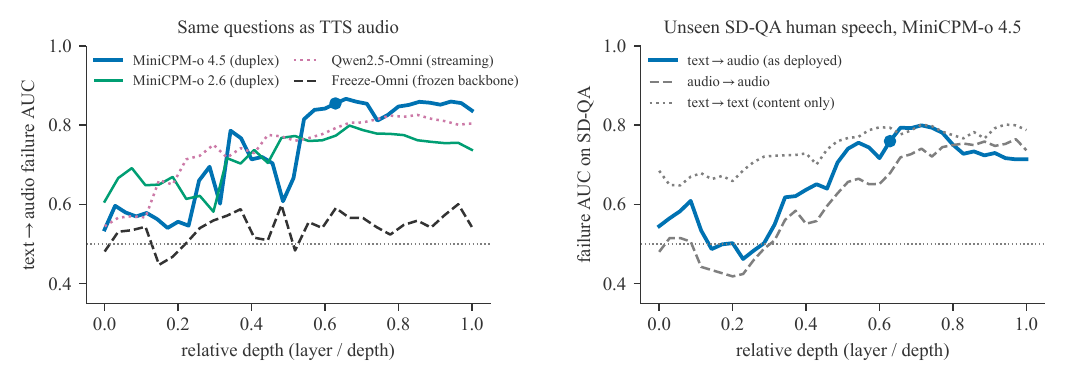}
  \caption{\textbf{Text-trained failure probes read speech states.}
  Left: text-trained probes scored on TTS audio of the same 600
  questions. Right: MiniCPM-o~4.5 on 200 unseen SD-QA human recordings,
  with audio-trained and text-input references in gray. Dots mark the
  deployed L22.}
  \label{fig:xmodal-transfer}
\end{figure}

\subsection{Probes trained on text read spoken questions}
\label{sec:finding-xmodal}
\label{sec:xmodal}
Correctness labels are cheap to collect for typed questions and costly
for recorded speech. We therefore ask whether a failure probe fitted on
text-input states still works when the same question is spoken. We
present 600 questions both as text and as TTS audio, train one probe per
layer on the text states, and score it on the audio states
(Figure~\ref{fig:xmodal-transfer}, left). For MiniCPM-o~4.5, the early
layers do not transfer: L0--L8 reach only 0.54--0.60 AUC. From L19 on,
about 55\% of the depth, the probe reads the audio states at
0.81--0.87, including 0.855 at the deployed L22. MiniCPM-o~2.6 and
Qwen2.5-Omni also rise with depth, to 0.80 and 0.83, while Freeze-Omni,
which keeps its language backbone frozen, stays near chance (at most
0.60). The transfer also holds for new questions and real voices: on
200 query-disjoint SD-QA human recordings, the text-trained probe
reaches 0.760 at L22 and 0.800 at L25 (right). Early layers are thus
modality-specific, while from mid-depth text and speech share the
direction that separates failures.

\subsection{The late-layer decline is specific to text input}
\label{sec:finding-cliff}
With text input, MiniCPM-o~4.5's last-token probe falls from 0.931 AUC
at L22 to 0.366 at the final layer, below chance
(Figure~\ref{fig:layersweep}). The raw Qwen3-8B backbone keeps 0.958
there, so MiniCPM-o's own training introduced the decline. It
does not appear when the same questions are spoken: fitted and tested
on audio input, the final layer reaches 0.936. Nor does it follow the
output mode, since text questions prefilled under the speak-mode
template still give 0.362. We conjecture that duplex training let turn
control take over the late layers on the text path. MiniCPM-o predicts
a binary Listen/Speak token before any content
\citep[\S3.3]{openbmb2026minicpmo}, and a weight intervention links the
two: moving L28--35 halfway back to the backbone raises L34 from 0.347
to 0.634, but the model then no longer ends its turns
(Appendix~\ref{app:mech}). Whether joint duplex pretraining or the
later SFT and RL stages \citep[\S5]{openbmb2026minicpmo} caused this
shift, and why audio input escapes it, remain open.

\section{Human Evaluation}
\label{sec:human-eval}
We compared live voice-only conversations with \emph{MiniCPM-o~4.5} using
\emph{always-local} and \emph{+ gate, aggressive}
(\emph{local} and \emph{aggressive} below).

\paragraph{Study setup.}
We report live evaluations of simple-fact (S1) and real-time-information (S2)
tasks: 16 task pairs (32 rated conversations) from 13 participants,
with 7 pairs in S1 and 9 in S2.
With model names hidden, participants could adapt prompts and finish early
within a nominal two-minute limit. They scored each conversation from 1 to 5 on
\emph{Correctness} (answer accuracy),
\emph{Task Helpfulness} (completing the task while following requirements and constraints),
\emph{Context Consistency} (remembering earlier information and staying consistent across turns), and
\emph{Conversation Naturalness} (appropriate timing, natural turn-taking, and clear, concise speech).
Each pair ended with a preference or tie and a reason.
On escalated turns, \emph{aggressive} spoke the holding line of
\S\ref{sec:system} and then waited silently for the relay, so its ratings,
including Naturalness, cover these waits.

\begin{figure}[t]
  \centering
  \includegraphics[width=\linewidth]{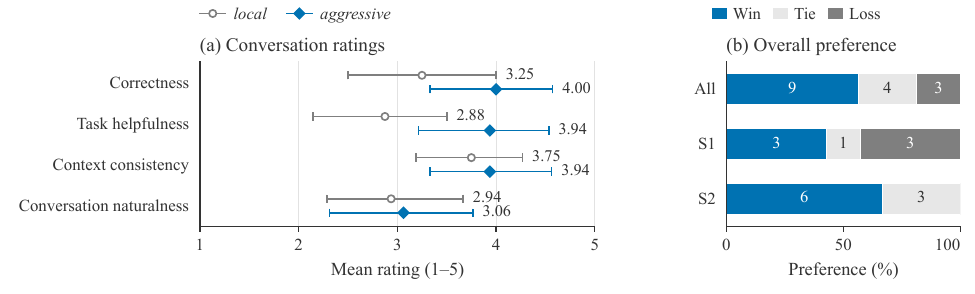}
  \caption{\textbf{Conversation ratings and preferences.}
  (a) Means and 95\% bootstrap intervals (10,000 participant resamples,
  16 conversations per policy). (b) \emph{Aggressive} wins/ties/losses.}
  \label{fig:human-eval}
\end{figure}

\paragraph{Better answers, especially for real-time information.}
Across S1 and S2, \emph{aggressive} received higher mean scores on all four
dimensions (Figure~\ref{fig:human-eval}a). The largest gains were in Correctness
(4.00 vs.\ 3.25) and Task Helpfulness (3.94 vs.\ 2.88), with smaller gains
in Context Consistency (3.94 vs.\ 3.75) and Conversation Naturalness
(3.06 vs.\ 2.94).
The benefit was clearest in S2: \emph{aggressive} scored higher on all four
dimensions, with Correctness rising from 2.56 to 3.67 and Task Helpfulness
from 2.00 to 3.67. It escalated on 19 of 37 recorded turns (51.4\%),
suggesting that access to the expert helped meet real-time information needs.
Users preferred it in 6 of 9 S2 pairs and tied the other 3.
Preference was more evenly split on simple facts (3 wins, 1 tie, 3 losses).
Across the two tasks, \emph{aggressive} received 9 wins, 4 ties, and 3 losses
(Figure~\ref{fig:human-eval}b). These preliminary results suggest that expert
access is especially useful for real-time information, but they do not establish
a broader conversational advantage.

\section{Discussion}
\label{sec:discussion}

\paragraph{Decision timing is part of routing quality.}
Useful failure information is accessible when the duplex model commits to
speak, so the routing decision can reuse computation already performed for
the conversation. This timing matters most on turns that remain local:
the transcript-router comparison shows that an additional model judgment
can delay these responses (\S\ref{sec:transfer}). Reading the native state
therefore provides a practical way to add competence arbitration while
preserving the existing interaction policy. The speech model manages the
floor, and the gate decides whether expert help is worth requesting.

\paragraph{Expert access as a deployment policy.}
The gains over matched random routing show that the readout directs expert
access toward turns where it is more useful. Keeping this decision outside
the generation policy makes the allocation adjustable: a serving threshold
can change the balance between local responses and expert calls without
updating the duplex checkpoint. Failure risk is a proxy for the value of
escalation, whose benefit also depends on the expert. This motivates
calibrating the operating point for the deployed local--expert pair and
its workload.

\paragraph{Waiting should serve the user's task.}
The small live study connects routing to task completion: expert access
was most helpful for real-time information, where a fluent local response
alone may not supply the facts the user needs (\S\ref{sec:human-eval}).
This suggests a useful design objective for speech agents: preserve fast
responses when local knowledge suffices and spend additional time when it
can improve the answer. Selective escalation gives an existing duplex
system a concrete way to make that choice at the response boundary.

\paragraph{A basis for adaptive competence control.}
Transfer to unseen human speech and gains from refitting on a second duplex
family in cached replay support the readout recipe beyond one input setting
or checkpoint. The read-time analysis offers a further direction: later
states improve aggregate failure prediction, while execution slips remain
harder to identify at onset (Appendix~\ref{app:readtime}). A natural extension
is to reassess risk as the answer develops, balancing additional evidence
against delayed speech or correction of content already spoken. This would
extend the current onset decision into competence monitoring during a turn.

\section{Conclusion}
\label{sec:conclusion}
We have shown that a frozen full-duplex speech model can support selective
expert handoff through a lightweight readout at its native listen-to-speak
transition. The gate reads existing answer-onset states without additional
language-model generation. Its threshold can be adjusted without repeating
duplex training. Across internal and external speech-QA evaluations, the
policy improves answer quality over local answering and matched random
routing. On the internal benchmark, it also reduces mean server waiting
relative to always escalating. The preliminary live
study further connects expert access to useful conversations, with the
clearest benefits on real-time-information tasks.

The broader contribution is a practical separation of interaction control
and competence arbitration. Representation analyses show that failure
information remains accessible across text and speech, including unseen
human recordings, and refitting on a second model family supports portability
of the training recipe in cached replay. Together, these results provide a
route to extending existing speech systems: retain the duplex model's
conversational policy, use its own states to guide expert access, and tune
the allocation as serving needs change. This division of labor lets fast
local interaction and stronger answers coexist within the same speech agent.
\label{sec:main-end}

\section*{AI use statement}

In this work, we used generative AI tools to implement methods (writing
and editing the experiment, serving, and analysis code) and to provide
feedback on the design of individual experiments. We have not used
generative AI tools to generate synthetic datasets, to develop theoretical models, or to
formulate or prove mathematical claims. Additionally, we used generative
AI tools to edit and polish parts of this manuscript, to produce
figure-generation scripts, and to search and summarize related
literature. Separately from authoring assistance, two commercial language
models are components of the system under study and are disclosed in the
paper itself: \texttt{gpt-5.4-mini} supplies answer-adequacy labels as
the training judge, and \texttt{gpt-5.5} serves as the escalation expert
(Section~\ref{sec:models}). We have reviewed all AI-assisted work:
AI-generated code was verified by the authors and is exercised by
regression tests, reported numbers were checked against the original
recorded data, and all claims and cited
references were checked against the underlying experiments and sources by
the authors. We take responsibility for the final content of this work,
including text, claims, and artifacts produced with the aid of generative
AI.

\section*{Ethics statement}

Alongside public-benchmark evaluation, Section~\ref{sec:human-eval} reports
a live user study that collected ratings, preferences, and interaction logs.
The paper artifacts contain aggregate rating, preference, and expert-use statistics.
Participant identifiers, recordings, and transcripts are not included.
The system itself raises one deployment
consideration: selective escalation transmits user audio, or text derived
from it, to a third-party cloud model. Deployments should disclose this
to users and handle the forwarded content in compliance with applicable
privacy requirements.

\section*{Reproducibility statement}

Models, evaluation pools, readout construction, and the live serving
protocol are specified in Section~\ref{sec:setup}, with per-experiment
detail in the appendix (Appendices~\ref{app:current-protocol}--\ref{app:readtime}).
Judge and self-evaluation prompts appear verbatim in
Appendix~\ref{app:prompts}, which also records seeds, software versions,
and hardware. The supplementary release includes the
calibration manifest, out-of-fold scores, thresholds, overlap exclusions,
benchmark hashes, and the scripts that rebuild the benchmark figures,
accuracy, and timing summaries from the original recorded data. For the
live study, we provide the statistics used in the paper and a figure-generation
script. Raw participant records are not part of these paper artifacts.

\clearpage
\setlength{\bibsep}{2.5pt} %
\bibliographystyle{iclr2027_conference}
\bibliography{refs}

@article{kadavath2022language,
  title   = {Language Models (Mostly) Know What They Know},
  author  = {Kadavath, Saurav and Conerly, Tom and Askell, Amanda and Henighan, Tom and Drain, Dawn and Perez, Ethan and Schiefer, Nicholas and Hatfield-Dodds, Zac and DasSarma, Nova and Tran-Johnson, Eli and others},
  journal = {arXiv preprint arXiv:2207.05221},
  year    = {2022}
}

@article{lin2022teaching,
  title   = {Teaching Models to Express Their Uncertainty in Words},
  author  = {Lin, Stephanie and Hilton, Jacob and Evans, Owain},
  journal = {Transactions on Machine Learning Research},
  year    = {2022},
  url     = {https://openreview.net/forum?id=8s8K2UZGTZ},
  note    = {arXiv:2205.14334}
}

@inproceedings{azaria2023internal,
  title     = {The Internal State of an {LLM} Knows When It's Lying},
  author    = {Azaria, Amos and Mitchell, Tom},
  booktitle = {Findings of the Association for Computational Linguistics: EMNLP 2023},
  year      = {2023},
  pages     = {967--976},
  publisher = {Association for Computational Linguistics},
  doi       = {10.18653/v1/2023.findings-emnlp.68},
  url       = {https://aclanthology.org/2023.findings-emnlp.68/},
  note      = {arXiv:2304.13734}
}

@inproceedings{burns2023discovering,
  title     = {Discovering Latent Knowledge in Language Models Without Supervision},
  author    = {Burns, Collin and Ye, Haotian and Klein, Dan and Steinhardt, Jacob},
  booktitle = {International Conference on Learning Representations (ICLR)},
  year      = {2023},
  note      = {arXiv:2212.03827}
}

@inproceedings{kuhn2023semantic,
  title     = {Semantic Uncertainty: Linguistic Invariances for Uncertainty Estimation in Natural Language Generation},
  author    = {Kuhn, Lorenz and Gal, Yarin and Farquhar, Sebastian},
  booktitle = {International Conference on Learning Representations (ICLR)},
  year      = {2023},
  note      = {arXiv:2302.09664}
}

@article{farquhar2024detecting,
  title   = {Detecting hallucinations in large language models using semantic entropy},
  author  = {Farquhar, Sebastian and Kossen, Jannik and Kuhn, Lorenz and Gal, Yarin},
  journal = {Nature},
  volume  = {630},
  pages   = {625--630},
  year    = {2024},
  doi     = {10.1038/s41586-024-07421-0},
  url     = {https://www.nature.com/articles/s41586-024-07421-0}
}

@inproceedings{orgad2025llms,
  title     = {{LLMs} Know More Than They Show: On the Intrinsic Representation of {LLM} Hallucinations},
  author    = {Orgad, Hadas and Toker, Michael and Gekhman, Zorik and Reichart, Roi and Szpektor, Idan and Kotek, Hadas and Belinkov, Yonatan},
  booktitle = {International Conference on Learning Representations (ICLR)},
  year      = {2025},
  url       = {https://arxiv.org/abs/2410.02707},
  note      = {arXiv:2410.02707}
}

@inproceedings{marks2024geometry,
  title     = {The Geometry of Truth: Emergent Linear Structure in Large Language Model Representations of True/False Datasets},
  author    = {Marks, Samuel and Tegmark, Max},
  booktitle = {Conference on Language Modeling (COLM)},
  year      = {2024},
  note      = {arXiv:2310.06824}
}

@article{alain2016understanding,
  title   = {Understanding intermediate layers using linear classifier probes},
  author  = {Alain, Guillaume and Bengio, Yoshua},
  journal = {arXiv preprint arXiv:1610.01644},
  year    = {2016}
}

@article{chen2023frugalgpt,
  title   = {{FrugalGPT}: How to Use Large Language Models While Reducing Cost and Improving Performance},
  author  = {Chen, Lingjiao and Zaharia, Matei and Zou, James},
  journal = {arXiv preprint arXiv:2305.05176},
  year    = {2023}
}

@article{ong2024routellm,
  title   = {{RouteLLM}: Learning to Route {LLMs} with Preference Data},
  author  = {Ong, Isaac and Almahairi, Amjad and Wu, Vincent and Chiang, Wei-Lin and Wu, Tianhao and Gonzalez, Joseph E. and Kadous, M. Waleed and Stoica, Ion},
  journal = {arXiv preprint arXiv:2406.18665},
  year    = {2024}
}

@inproceedings{ding2024hybrid,
  title     = {Hybrid {LLM}: Cost-Efficient and Quality-Aware Query Routing},
  author    = {Ding, Dujian and Mallick, Ankur and Wang, Chi and Sim, Robert and Mukherjee, Subhabrata and R{\"u}hle, Victor and Lakshmanan, Laks V.~S. and Awadallah, Ahmed Hassan},
  booktitle = {International Conference on Learning Representations (ICLR)},
  year      = {2024},
  note      = {arXiv:2404.14618}
}

@article{defossez2024moshi,
  title   = {Moshi: a speech-text foundation model for real-time dialogue},
  author  = {D{\'e}fossez, Alexandre and Mazar{\'e}, Laurent and Orsini, Manu and Royer, Am{\'e}lie and P{\'e}rez, Patrick and J{\'e}gou, Herv{\'e} and Grave, Edouard and Zeghidour, Neil},
  journal = {arXiv preprint arXiv:2410.00037},
  year    = {2024}
}

@article{wang2024freezeomni,
  title   = {{Freeze-Omni}: A Smart and Low Latency Speech-to-speech Dialogue Model with Frozen {LLM}},
  author  = {Wang, Xiong and Li, Yangze and Fu, Chaoyou and Shen, Yunhang and Xie, Lei and Li, Ke and Sun, Xing and Ma, Long},
  journal = {arXiv preprint arXiv:2411.00774},
  year    = {2024}
}

@inproceedings{wang2024fsm,
  title     = {A Full-duplex Speech Dialogue Scheme Based On Large Language Model},
  author    = {Wang, Peng and Lu, Songshuo and Tang, Yaohua and Yan, Sijie and Xia, Wei and Xiong, Yuanjun},
  booktitle = {Advances in Neural Information Processing Systems (NeurIPS)},
  year      = {2024},
  volume    = {37},
  pages     = {13372--13403},
  doi       = {10.52202/079017-0427},
  note      = {arXiv:2405.19487}
}

@inproceedings{zhang2024duplex,
  title     = {Beyond the Turn-Based Game: Enabling Real-Time Conversations with Duplex Models},
  author    = {Zhang, Xinrong and Chen, Yingfa and Hu, Shengding and Han, Xu and Xu, Zihang and Xu, Yuanwei and Zhao, Weilin and Sun, Maosong and Liu, Zhiyuan},
  booktitle = {Proceedings of the 2024 Conference on Empirical Methods in Natural Language Processing},
  year      = {2024},
  pages     = {11543--11557},
  publisher = {Association for Computational Linguistics},
  doi       = {10.18653/v1/2024.emnlp-main.644},
  url       = {https://aclanthology.org/2024.emnlp-main.644/},
  note      = {arXiv:2406.15718}
}

@inproceedings{veluri2024syncllm,
  title     = {Beyond Turn-Based Interfaces: Synchronous {LLMs} as Full-Duplex Dialogue Agents},
  author    = {Veluri, Bandhav and Peloquin, Benjamin N. and Yu, Bokai and Gong, Hongyu and Gollakota, Shyamnath},
  booktitle = {Proceedings of the 2024 Conference on Empirical Methods in Natural Language Processing},
  year      = {2024},
  pages     = {21390--21402},
  publisher = {Association for Computational Linguistics},
  doi       = {10.18653/v1/2024.emnlp-main.1192},
  url       = {https://aclanthology.org/2024.emnlp-main.1192/},
  note      = {arXiv:2409.15594}
}

@article{huang2026duplexomni,
  title   = {{DuplexOmni}: Real-Time Listening, Seeing, Thinking, and Speaking for Full-Duplex Interaction},
  author  = {Huang, Muye and Zhang, Lingling and Yu, Xingyu and Shi, Lei and Ma, Zhanyu and Xu, Jun and Gao, Jiuchong and Hao, Jinghua and He, Renqing and Liu, Jun},
  journal = {arXiv preprint arXiv:2606.09186},
  year    = {2026}
}

@inproceedings{chien2026moshirag,
  title     = {{MoshiRAG}: Asynchronous Knowledge Retrieval for Full-Duplex Speech Language Models},
  author    = {Chien, Chung-Ming and Orsini, Manu and Kharitonov, Eugene and Zeghidour, Neil and Livescu, Karen and D{\'e}fossez, Alexandre},
  booktitle = {International Conference on Machine Learning (ICML)},
  year      = {2026},
  note      = {arXiv:2604.12928}
}

@article{openbmb2026minicpmo,
  title   = {{MiniCPM-o 4.5}: Towards Real-Time Full-Duplex Omni-Modal Interaction},
  author  = {Cui, Junbo and Xu, Bokai and Wang, Chongyi and Yu, Tianyu and Sun, Weiyue and Xu, Yingjing and others},
  journal = {arXiv preprint arXiv:2604.27393},
  year    = {2026}
}

@article{chen2024voicebench,
  title   = {{VoiceBench}: Benchmarking {LLM}-Based Voice Assistants},
  author  = {Chen, Yiming and Yue, Xianghu and Zhang, Chen and Gao, Xiaoxue and Tan, Robby T. and Li, Haizhou},
  journal = {arXiv preprint arXiv:2410.17196},
  year    = {2024}
}

@inproceedings{faisal2021sdqa,
  title     = {{SD-QA}: Spoken Dialectal Question Answering for the Real World},
  author    = {Faisal, Fahim and Keshava, Sharlina and Alam, Md Mahfuz ibn and Anastasopoulos, Antonios},
  booktitle = {Findings of the Association for Computational Linguistics: EMNLP 2021},
  year      = {2021},
  pages     = {3296--3315},
  publisher = {Association for Computational Linguistics},
  doi       = {10.18653/v1/2021.findings-emnlp.281},
  url       = {https://aclanthology.org/2021.findings-emnlp.281/},
  note      = {arXiv:2109.12072}
}

@article{kossen2024semantic,
  title   = {Semantic Entropy Probes: Robust and Cheap Hallucination Detection in {LLMs}},
  author  = {Kossen, Jannik and Han, Jiatong and Razzak, Muhammed and Schut, Lisa and Malik, Shreshth and Gal, Yarin},
  journal = {arXiv preprint arXiv:2406.15927},
  year    = {2024}
}

@article{bayling2026duplex,
  title   = {{BayLing-Duplex}: Native Full-Duplex Speech Dialogue with a Single Autoregressive {LLM}},
  author  = {Fang, Qingkai and Guo, Shoutao and Feng, Yang},
  journal = {arXiv preprint arXiv:2606.14528},
  year    = {2026}
}

@article{zhang2026liberating,
  title   = {Liberating {LLM} Capabilities in Full-Duplex Speech Models},
  author  = {Zhang, Luoyuan and Xu, Bokai and Cui, Junbo and Sun, Weiyue and Xu, Yingjing and Liu, Hanyu and Yao, Yuan},
  journal = {arXiv preprint arXiv:2606.07547},
  year    = {2026}
}

@inproceedings{chrono2026thinking,
  title     = {Chronological Thinking in Full-Duplex Spoken Dialogue Language Models},
  author    = {Wu, Donghang and Zhang, Haoyang and Chen, Chen and Zhang, Tianyu and Tian, Fei and Yang, Xuerui and Yu, Gang and Liu, Hexin and Hou, Nana and Hu, Yuchen and Chng, Eng Siong},
  booktitle = {Proceedings of the 27th Annual Meeting of the Special Interest Group on Discourse and Dialogue (SIGDIAL)},
  year      = {2026},
  pages     = {473--485},
  publisher = {Association for Computational Linguistics},
  doi       = {10.18653/v1/2026.sigdial-1.33},
  url       = {https://aclanthology.org/2026.sigdial-1.33/},
  note      = {arXiv:2510.05150}
}

@article{li2025baichuanaudio,
  title   = {{Baichuan-Audio}: A Unified Framework for End-to-End Speech Interaction},
  author  = {Li, Tianpeng and Liu, Jun and Zhang, Tao and Fang, Yuanbo and Pan, Da and Wang, Mingrui and others},
  journal = {arXiv preprint arXiv:2502.17239},
  year    = {2025},
  note    = {Introduces OpenAudioBench}
}

@inproceedings{mahaut2024factual,
  title     = {Factual Confidence of {LLM}s: on Reliability and Robustness of Current Estimators},
  author    = {Mahaut, Mat{\'e}o and Aina, Laura and Czarnowska, Paula and Hardalov, Momchil and M{\"u}ller, Thomas and Marquez, Lluis},
  booktitle = {Proceedings of the 62nd Annual Meeting of the Association for Computational Linguistics (Volume 1: Long Papers)},
  pages     = {4554--4570},
  address   = {Bangkok, Thailand},
  publisher = {Association for Computational Linguistics},
  doi       = {10.18653/v1/2024.acl-long.250},
  year      = {2024}
}

@article{lugoloobi2026failures,
  title   = {{LLM}s Encode Their Failures: Predicting Success from Pre-Generation Activations},
  author  = {Lugoloobi, William and Foster, Thomas and Bankes, William and Russell, Chris},
  journal = {arXiv preprint arXiv:2602.09924},
  year    = {2026}
}

@article{varshney2026llmrouter,
  title   = {{LLM} Router: Rethinking Routing with Prefill Activations},
  author  = {Varshney, Tanay and Surla, Annie and Xu, Michelle and Krishnan, Gomathy Venkata and Jeblick, Maximilian and Austin, David and Vaidya, Neal and Onofrio, Davide},
  journal = {arXiv preprint arXiv:2603.20895},
  year    = {2026}
}

@misc{sglangomniinteractive2026,
  title        = {Unified Runtime for Full-Duplex Interactive Intelligence with {SGLang Omni}},
  author       = {{SGLang-Omni Team}},
  howpublished = {\url{https://www.sglang.io/blog/omni-interactive-runtime}},
  year         = {2026},
  month        = oct
}

\clearpage
\appendix
\section{Evaluation and calibration details}
\label{app:current-protocol}

The Internal test set contains 240 queries: chat (60), factual QA (40),
knowledge (60), math (60), and SimpleQA traps (20). Each query has unit
weight, and every original query remains in the accuracy denominator. The four
external English pools retain their original membership.

\subsection{Experimental setup details}
\label{app:setup-details}

\paragraph{Models and decoding.}
The target is MiniCPM-o~4.5 (9B, based on Qwen3-8B). Every experiment,
including the live study, loads it without its vision encoder
(\texttt{init\_vision=False}) and sends only audio or text, never images or
video. Controls include
Qwen3-8B, MiniCPM-o~2.6, Qwen2.5-7B, Qwen2.5-Omni, Qwen2-Audio, and
Freeze-Omni. Representation analyses use bf16, greedy decoding, and seed 42.
Native decoding uses temperature $T=0.7$ and top-$k=20$. The expert is
\texttt{gpt-5.5} with low reasoning effort and the web-search tool.

\paragraph{Probe fitting and thresholds.}
Representation probes use 360 calibration queries and single-position
4,096-dimensional features. The native gate uses 5,228 calibration rows
and three L22 summaries: the last tail state, the tail mean, and the
input mean. Its logistic head uses $C=3\times10^{-4}$.
Five-fold stratified cross-validation uses shuffle seed 42.
For each language, conservative, balanced, and aggressive thresholds
are the $(1-r)$ quantiles of the 4,986 core rows' out-of-fold scores,
with $r\in\{0.15,0.30,0.50\}$. The 242 fresh training rows enter
the fit but not the quantile base. Every reported live run and the
baselines of Appendix~\ref{app:textrouter} use these 5,228 rows.

\paragraph{Judge configuration.}
Speech TriviaQA, Speech Web Questions, and Speech Llama Questions use
the OpenAudioBench judge with \texttt{gpt-4o-2024-08-06}. Internal,
SD-QA, and Mandarin reasoning answers use the reference-anchored
\texttt{gpt-5.4-mini} adequacy judge. The adequacy judge
receives the question, an optional reference, and a candidate answer
without the system identity. Its rubric requires factual correctness,
valid reasoning, and a complete answer. Prompt details are in
Appendix~\ref{app:prompts}.

\subsection{Metric provenance}
\label{app:metric-sources}
In the main run, accuracy, escalation rates, and TTFA come from the same recorded
sessions, and each arm is scored on its own delivered answers: one
attempt per query and policy, inputs replayed in real
time, expert requests built from the full query audio (transcription
begins at input end and never before the last sample's
scheduled arrival), and both local and relay paths producing audio. The
delivered answer of each session is judged with the per-pool judges of
Table~\ref{tab:transfer}. Failed sessions stay in the denominator and
count as incorrect. Routing comparisons against matched random are computed within
one arm set, so all arms share the same expert pipeline. TTFA is available for 5,903 of 5,950 attempts, and 61
completed early responses contribute zero wait. The holding line of
Section~\ref{sec:system} and client playback are outside the endpoint. The
benchmark sessions give the holding line to the duplex model as a text note
instead of playing the pre-synthesized audio, and neither metric uses that
audio.
Specifically, reported TTFA is
$\max(0,t_{\rm first\ answer\ PCM}-t_{\rm scheduled\ input\ end})$
on the server's monotonic clock. Local timing ends when the first
PCM-bearing generation call returns; relay timing ends at the first
yielded answer PCM block. All completed waits are retained in the main
results. Appendix~\ref{app:llmdeploy} reports the separate deployment
runs in Figure~\ref{fig:dualview}'s bottom row and an explicitly
filtered sensitivity analysis. The scheduled input end is available to
both routing methods in this benchmark; endpoint-detection delay in
live speech and client-audible latency are not measured.

\subsection{TTFA decomposition and serving-pipeline overhead}
\label{app:ttfa-decomp}
Table~\ref{tab:ttfa-decomp} splits each arm's mean TTFA in the main run
into the expert's own response time (the reasoning-plus-search latency
of the cloud model, zero on unescalated turns) and everything the
serving system adds around it: waiting for the local commit,
transcribing the uplink, synthesizing the relay, and scheduling. The
pipeline transcribes the full query as soon as its last sample has
arrived (never earlier, since the transcription is input-causal), fires
the always arm at input end rather than at the local commit, and
delivers the relay the moment the expert answer is ready. The remaining
overhead ($\approx$1.7--2.4\,s mean) is dominated by the portion of
transcription not hidden behind the commit wait and by the first relay
synthesis block.

\begin{table}[h]
\centering\small
\caption{Mean server TTFA in the main run decomposed as expert
response time plus system overhead (gate wait, transcription, relay
synthesis, scheduling), over all completed sessions (nonnegative-wait
convention, with zero expert time for unfired sessions). On the always
arm, where every session escalates, the system overhead beyond the
expert call is 2.39\,s mean / 1.96\,s median on Internal
(external pools pooled: 1.85/1.67).}
\label{tab:ttfa-decomp}
\begin{tabular}{lcc}
\toprule
tier & Internal (TTFA = expert + system) & External macro \\
\midrule
local-only & 1.08 = 0.00 + 1.08 & 1.74 = 0.00 + 1.74 \\
conservative & 2.15 = 0.85 + 1.31 & 1.75 = 0.27 + 1.48 \\
balanced & 4.67 = 2.96 + 1.71 & 2.45 = 0.87 + 1.59 \\
aggressive & 5.95 = 3.92 + 2.03 & 4.07 = 2.01 + 2.06 \\
always & 8.68 = 6.28 + 2.39 & 5.38 = 3.54 + 1.85 \\
\bottomrule
\end{tabular}
\end{table}

\subsection{Internal task strata}
\label{app:internal-strata}
\begin{table}[h]
\centering\small
\caption{Descriptive strata of the full 240-query Internal test set, with unit query weights. Local, aggressive (A), and always report answer-content accuracy (\%), and call rate is the aggressive arm's realized rate. All five categories and all original queries are retained.}
\label{tab:internal-strata}
\begin{tabular}{lrrrrr}
\toprule
Stratum & $n$ & Local & A & Always & A call rate \\
\midrule
Chat & 60 & 61.7 & 70.0 & 80.0 & 30.0 \\
Factual QA & 40 & 42.5 & 92.5 & 95.0 & 60.0 \\
Knowledge & 60 & 13.3 & 45.0 & 63.3 & 71.7 \\
Math & 60 & 65.0 & 66.7 & 80.0 & 25.0 \\
SimpleQA traps & 20 & 0.0 & 100.0 & 95.0 & 100.0 \\
\bottomrule
\end{tabular}
\end{table}

The largest gains occur on factual QA and SimpleQA traps. The fixed mix is not a
production estimate, but it preserves the full composition and every category.

The Internal math slice illustrates a threshold limitation. Of the five
GSM8K failures the expert can fix, two restate the problem without
solving it and three contain an arithmetic or setup error. Their
readout scores rank above most GSM8K successes (AUC 0.82) but all fall
below the aggressive threshold (at most 0.50 vs.\ 0.55), so the miss is
the operating point rather than the ranking.

\section{Representation and readout details}

\subsection{Layer and modality controls}
\label{app:signal}
\paragraph{Text-input transfer.}
We evaluate the readout without using expert outcomes. The sweep in
Figure~\ref{fig:layersweep} compares read positions and
raw-backbone controls, and mean pooling reduces late-layer degradation.
A raw Qwen backbone, subsampled to match per-pool failure rates,
remains near 0.96, with similar ordering in the Qwen2.5 control
family (Table~\ref{tab:lopo}).
With audio calibration and audio input, the final-layer probe reaches
0.936 on LOPO math. These controls link degradation to the tested
speech checkpoints and input regimes without identifying its cause
(Appendix~\ref{app:mech}).

Table~\ref{tab:lopo} compares the duplex checkpoints with their raw backbones
under the text-input LOPO-math protocol. Mid-layer transfer persists in both
duplex models, whereas the final-layer decline is absent or smaller in the raw
and streaming-omni controls.

\begin{table}[h]
\centering\small
\caption{Text-input LOPO math AUC by depth for matched model families.}
\label{tab:lopo}
\begin{tabular}{lccl}
\toprule
model & best mid-layer & final layer & depth profile \\
\midrule
Qwen3-8B & 0.964 (L33/36) & 0.958 & plateau \\
MiniCPM-o 4.5 & \textbf{0.931 (L22/36)} & \textbf{0.366} & late decline \\
Qwen2.5-7B & 0.893 (L21/28) & 0.809 & mild dip \\
Qwen2.5-Omni & 0.794 (L16/28) & 0.746 & diffuse decline \\
MiniCPM-o 2.6 & \textbf{0.822 (L21/28)} & \textbf{0.540} & late decline \\
\bottomrule
\end{tabular}
\end{table}

\subsection{Mechanism checks}
\label{app:mech}
A speak-mode prompt template leaves final-layer LOPO math essentially unchanged
(0.362 versus 0.366), the audio--text direction is nearly orthogonal to the
failing probe, and source-pool identity remains decodable at all depths.

\paragraph{Weight interpolation.} MiniCPM-o~4.5 keeps the Qwen3-8B layer
tensors (397 of 399 match in name and shape, and only the embedding and output head
differ). We interpolate the L28--35 weights toward the backbone,
$w\leftarrow(1-\beta)\,w_{\text{MiniCPM}}+\beta\,w_{\text{Qwen}}$, repeat the
text-input LOPO math sweep, and replay 100 native duplex queries
(Table~\ref{tab:backdiff}). Every model answers all 100 queries, and L22 stays
at 0.931 because earlier layers are unchanged. The late-layer readout recovers
with the dose. At $\beta=0.5$ the model answers but never ends its turn.
Reverting only the MLPs recovers L33 as much as the full change without lost
turn endings, so readout damage and turn ending do not sit in exactly the same
weights. The intervention links the late-layer change to turn behavior, but it does
not identify the training stage or objective responsible.

\begin{table}[h]
\centering\small
\caption{Weight interpolation of L28--35 toward Qwen3-8B: text-input LOPO math
AUC of last-token probes and turns ended in 100 native duplex queries.}
\label{tab:backdiff}
\begin{tabular}{llrrrr}
\toprule
$\beta$ & modules & L33 & L34 & L35 & turns ended \\
\midrule
0 & -- & 0.498 & 0.347 & 0.364 & 99/100 \\
0.25 & attention only & 0.492 & 0.401 & 0.423 & 97/100 \\
0.25 & MLP only & 0.591 & 0.477 & 0.382 & 100/100 \\
0.25 & all & 0.591 & 0.515 & 0.434 & 87/100 \\
0.5 & all & 0.660 & 0.634 & 0.435 & 0/100 \\
\bottomrule
\end{tabular}
\end{table}

\section{Portability to a second duplex family}
\label{app:nvda}
We replay NemotronLabs-VoiceChat-11B through its native frame protocol with the
checkpoint frozen. Standardized logistic heads read L26/L30/L34 onset states,
and their normalized logits are averaged. The recipe is selected using
calibration-only LOPO metrics. Its three-layer read reaches 0.818 out-of-fold
AUC and 0.816 mean external AUC. The read covers the first 640\,ms after commit,
so this experiment tests method portability rather than end-to-end timing.

\begin{table}[h]
\centering\small
\caption{VoiceChat-11B cached native-frame replay, equally averaged over four
external English speech-QA pools. Random uses the same call rate.}
\label{tab:nvda-summary}
\begin{tabular}{lrrr}
\toprule
expert budget & 15\% & 30\% & 50\% \\
\midrule
probe-guided accuracy & 55.0 & 65.0 & 75.7 \\
matched-random accuracy & 50.7 & 57.7 & 67.0 \\
\bottomrule
\end{tabular}
\end{table}

On the 240-query Internal replay, 223 queries yield a valid onset read, and the
remaining 17 stay local. At nominal 15/30/50\% scoreable-query budgets, the
policy attains 31.7/44.6/60.4\% accuracy, from a 19.6\% local endpoint, while
matched random reaches 29.6/39.9/53.5\%. The layer set was reviewed using the
Internal split, so we use the external LOPO result as the cleaner portability
evidence.

\section{Gate operation and detailed results}
\label{app:gatedetail}
\label{app:livedetail}

The L22 hidden state is available after roughly 57--61\% of prefill. On one
H100, probe availability has P50/P95 of 20/25\,ms for text and 45/104\,ms for
audio. These measurements cover the local read and exclude both the network
transfer and the expert-model call.

\paragraph{Dialogue-act filtering.}
\label{app:actgate}
A second linear head reads the same L22 state and scores whether the
turn is an information request rather than floor management, such as an
acknowledgment or a stop command. The gate escalates only turns that
clear both this threshold and the failure threshold. It classifies
every benchmark session it scores as an information request, so it
changes no reported benchmark result. It acts in live conversation.

\subsection{Realized expert-call rates}
\label{app:transferdetail}
\begin{table}[h]
\centering\small
\caption{Realized escalation rates (\%) for the MiniCPM results in Table~\ref{tab:transfer}. Internal retains all 240 queries, and every pool uses unit query weights. Rates can differ from nominal calibration and VoiceChat replay budgets.}
\label{tab:native-rates}
\begin{tabular}{lrrrrr}
\toprule
Pool & $n$ & Conservative & Balanced & Aggressive & Always \\
\midrule
Internal & 240 & 8.3 & 28.3 & 50.0 & 100.0 \\
Speech TriviaQA & 250 & 3.6 & 19.2 & 55.2 & 100.0 \\
Speech WebQ & 250 & 6.4 & 23.6 & 56.8 & 100.0 \\
Speech Llama Q. & 250 & 0.0 & 2.8 & 21.2 & 100.0 \\
SD-QA & 200 & 7.5 & 24.5 & 62.5 & 100.0 \\
\bottomrule
\end{tabular}
\end{table}

\subsection{Expert dollar cost}
\label{app:expertcost}
The benchmark run did not persist API billing usage, so the
USD/query row of Table~\ref{tab:transfer} is an estimate built from a
replay. Escalated uplink transcripts of the benchmark queries are
re-sent through the identical expert configuration (the same model, low
reasoning effort, web-search tool, and system prompt as the benchmark),
and the billed usage is priced at the provider's published rates
(input \$5/M tokens, cached input \$0.5/M, output \$30/M including
hidden reasoning tokens, web search \$10/1k calls, checked
2026-09-10). A pool$\times$arm entry is that cell's replayed mean cost
per escalation times the run's escalation count (2{,}059 in total),
over the pool size. Non-escalated queries cost \$0, and the external entry is
the equal-pool mean, matching the escalation row. The mean cost per
escalation is \$0.018: a replayed call bills on average 5{,}264 input
tokens (3{,}712 cached), dominated by the web-search tool definition
rather than the short transcripts, against 106 output tokens, and
12.7\% of calls invoke web search. Replay token counts vary with
decoding stochasticity and cache state. Costs exclude local compute,
ASR, and TTS. Sources, per-cell unit costs, and tallies are pinned in
\texttt{revision\_data/expert\_cost\_usd.json}.

\clearpage
\subsection{Internal accuracy and latency}
\begin{table}[h]
\centering\small
\caption{Internal results on all 240 original test queries (\%), with unit query weights. Random mixes the measured local/always endpoints at the corresponding call rate.}
\label{tab:live}
\begin{tabular}{lrrr}
\toprule
tier & calls & accuracy & random \\
\midrule
local-only & 0.0 & 42.1 & --- \\
\gaterow conservative & 8.3 & 48.8 & 45.2 \\
\gaterow balanced & 28.3 & 60.0 & 52.7 \\
\gaterow aggressive & 50.0 & 69.2 & 60.8 \\
always & 100.0 & 79.6 & --- \\
\bottomrule
\end{tabular}
\end{table}

\begin{table}[!ht]
\centering\small
\caption{Internal server TTFA (seconds), measured on all 240 test IDs: 240 attempts per arm, with 238/238/239/238/239 completed and 2/2/1/2/1 failed in row order. TTFA is the nonnegative wait from scheduled input end to first answer PCM. All completed sessions count, and the 19/16/14/9/0 early responses have zero wait. Failed attempts have no TTFA value. Timing ends at the first local PCM-bearing chunk return or relay PCM yield, excluding candidate/stall audio and client playback.}
\label{tab:latency}
\begin{tabular}{lrrrr}
\toprule
tier & mean & P50 & P95 & P99 \\
\midrule
local-only & 1.08 & 0.78 & 2.61 & 3.66 \\
\gaterow conservative & 2.15 & 0.92 & 7.23 & 22.74 \\
\gaterow balanced & 4.67 & 1.60 & 17.63 & 39.40 \\
\gaterow aggressive & 5.95 & 3.52 & 19.19 & 40.48 \\
always & 8.68 & 6.75 & 22.90 & 34.11 \\
\bottomrule
\end{tabular}
\end{table}

\subsection{Failure type and read time}
\label{app:readtime}
A separate sweep fits the same readout family at successive native read points.
Its five-pool macro includes Mandarin reasoning. Specific-but-wrong answers
account for about 70\% of failures and are ranked at AUC 0.813. Perception
errors reach 0.889, while execution slips reach 0.469. Reading just
before onset gives nearly the same external macro AUC as reading at
onset (0.753 vs.\ 0.755), so onset is the read point because the
native model commits to respond there, not because it carries extra
signal. Reading two answer chunks later raises external macro AUC to
0.791 (0.801 after three), at the cost of a later handoff.

\section{Routing baselines and comparisons}
\label{app:baselines}

\subsection{Gate versus matched random routing}
\label{app:pairedci}
Because each benchmark arm is an independently sampled run, arm-to-arm
accuracy differences mix routing effects with decoding and judging
noise. We therefore report the paired per-query difference
$\Delta(q)=\mathrm{ok}_{\text{gate}}(q)-[(1-r)\,\mathrm{ok}_{\text{local}}(q)+r\,\mathrm{ok}_{\text{always}}(q)]$,
where $r$ is the arm's realized escalation rate (Table~\ref{tab:pairedci}). As in the main
table, every arm is scored on its own delivered answers. At
the aggressive tier the external macro gain over matched random is
$+7.1$ points. A fixed-outcome replay that
holds each query's local-only and always-expert answers fixed and lets
the readout logged in the local-only session choose which to escalate
gives a similar estimate, $+6.9$ points. Conservative-tier gains are smaller,
at $+0.7$ points on the external macro and $+3.5$ points on Internal,
with a $\sim$4--8\% call rate. Per-cell values are recorded in
\texttt{data/paired\_delta\_ci\_ttfa2.json} (live) and
\texttt{data/ttfa2\_unified\_offline.json} (fixed outcomes).

\begin{table}[h]
\centering\small
\caption{Gate accuracy minus matched-random accuracy (points).
Live: independently sampled Table~\ref{tab:transfer}
arms, paired at the query level. Fixed outcomes: the local-only and
always-expert answers held fixed per query, escalation chosen by the
readout of the local-only session at the deployed thresholds.}
\label{tab:pairedci}
\begin{tabular}{lrr}
\toprule
Aggressive tier & Live & Fixed outcomes \\
\midrule
Internal & $+8.3$ & $+8.1$ \\
S.\ TriviaQA & $+10.0$ & $+8.8$ \\
S.\ WebQ & $+5.6$ & $+5.1$ \\
S.\ Llama Q. & $+4.5$ & $+4.7$ \\
SD-QA & $+8.4$ & $+8.9$ \\
External macro & $+7.1$ & $+6.9$ \\
\bottomrule
\end{tabular}
\end{table}

\subsection{Text-router baselines}
\label{app:textrouter}
We compare the deployed readout against alternatives trained on the
same 5{,}228 calibration rows with the same folds and labels
(Table~\ref{tab:textrouter}). On the external pools every score
ranks the main run's queries and is judged on its local-only and
always-expert answers, and the readout is the score logged at the
commit of the local-only session.
A TF--IDF logistic router on the gold question text matches
in-mix but collapses under pool transfer (En-4 mean AUC 0.53 against
0.78 for the readout). It reads the \emph{gold}
transcript (an upper bound unavailable at answer onset, where a
live system would add streaming-ASR latency and errors), while the
readout reuses hidden states already computed at no marginal cost.

\paragraph{A fast LLM as the router.} We also test the text classifier
a practitioner would reach for first: a fast LLM asked, zero-shot, to
rate the question. \texttt{gpt-5.4-mini} at its fastest setting
(reasoning effort \emph{none}) returns in one call three integer
probabilities: whether ``a small on-device voice assistant model,
answering from memory without tools, [would] likely get the answer to
the question wrong'', whether the question needs a lookup, and whether
it needs multi-step reasoning. We use the first rating as the router's score, as is, with nothing
fitted on our labels, and call the LLM used this way the zero-shot
router. Calibration rows have only gold text, while the external pools
use the live pipeline's ASR transcript, which is what a deployed text
router would see. On the external pools the zero-shot router ranks
failures worse than the readout, though not on every pool. Its En-4
AUC is 0.722 against 0.783 ($-0.061$), and its five-pool routing gain is
$+4.6$ against $+7.2$ points ($-2.6$). In the calibration mix it is
also weaker (AUC 0.767 vs.\ 0.848), and on Chinese rows much weaker
(0.578 vs.\ 0.782). Hidden states at the model's own commit point
therefore carry failure signal that the LLM's rating of the question
lacks. Appendix~\ref{app:llmdeploy} measures the router live and what
it costs when it replaces the readout. The LLM's training data are unknown and may include these public
benchmarks.

\begin{table}[h]
\centering\small
\caption{Failure-prediction AUC and routing gain over matched random
at the aggressive call rate (points, mean over the five pools), same
training rows, folds, and labels for all learned rows. On the external
pools every row ranks the main run's queries and is judged on its
local-only and always-expert answers. The ASR row scores the live
router's transcript, and calibration rows have only gold text. The
GPT-5.4 mini rows are unfitted, so their first column is plain AUC on
the gold calibration text.}
\label{tab:textrouter}
\begin{tabular}{lrrr}
\toprule
Score & OOF & En-4 AUC & Agg.\ gain (5-pool mean) \\
\midrule
TF--IDF, gold text & 0.764 & 0.532 & $+3.1$ \\
GPT-5.4 mini zero-shot, gold text & 0.767 & 0.723 & $+4.5$ \\
GPT-5.4 mini zero-shot, ASR text & 0.767 & 0.722 & $+4.6$ \\
Hidden-state readout (deployed) & 0.848 & 0.783 & $+7.2$ \\
\bottomrule
\end{tabular}
\end{table}

\paragraph{Where each signal is stronger.} The two scores answer
different questions. The zero-shot router is asked whether a small
assistant model would likely get the question wrong, so it rates the question from its
wording. The readout is trained on this model's own failures and reads
its hidden state at the commit. Splitting the calibration rows by
question family (Table~\ref{tab:probellmpools}), the largest margins favoring
the readout occur on long-tail entity questions from
PopQA (0.863 vs.\ 0.701, difference $+0.162$) and on the Chinese sets (all Chinese rows 0.782
vs.\ 0.578). The hidden state thus carries signal about this model's
failures that the wording does not show. How much of it reflects the
question rather than the state of one session is not separated here.
The router here is unfitted, while the readout scores are out-of-fold but
come from training on the same families.

\begin{table}[!ht]
\centering\small
\caption{Failure-prediction AUC by question family on the calibration
rows (gold question text, label = native failure of the local model).
Readout = deployed out-of-fold score, LLM = zero-shot
\texttt{gpt-5.4-mini} rating. Of the 17 families, PopQA and TruthfulQA (the only one
where the LLM scores higher) are shown. All 17 are in
\texttt{revision\_data/probe\_llm\_pools.json}, and one further family
with a single outcome class is omitted.}
\label{tab:probellmpools}
\begin{tabular}{lrrrr}
\toprule
Question family (source) & $n$ & Readout & LLM & Readout $-$ LLM \\
\midrule
Long-tail entities (PopQA) & 741 & 0.863 & 0.701 & +0.162 \\
TruthfulQA & 441 & 0.693 & 0.698 & $-$0.005 \\
\midrule
All English rows & 4828 & 0.848 & 0.771 & +0.077 \\
All Chinese rows & 400 & 0.782 & 0.578 & +0.204 \\
\bottomrule
\end{tabular}

\end{table}

\subsection{Confidence and head-form baselines}
\label{app:confbaselines}
Two cheaper alternatives read the same model at the same moment. Both are
measured on separate native feature-dump runs of the same queries, with
the deployed readout scored on the identical rows as the reference, so
neither enters Table~\ref{tab:transfer}.

\paragraph{Next-token confidence.} At the read point the language-model
head already gives a next-token distribution at no extra cost. Its entropy
and top-1 probability rank local failures near chance on every pool (AUC
0.51--0.58), whether read at the commit or just before it. The English
external mean is 0.545 at the commit and 0.519 just before it, against
0.775 for the deployed readout on the same rows. On Internal the three
scores are 0.506, 0.513, and 0.845.

\paragraph{Head form.} On the 5{,}228 calibration rows with the same
folds and labels, we replace the deployed head, a logistic regression
with $C=3\times10^{-4}$ on raw features, by four alternatives
(Table~\ref{tab:headform}). None improves the English external mean.
Elastic net is close to the deployed head. Per-dimension standardization
and both non-linear heads lose 0.06 to 0.07 of English external AUC.
Standardization alone costs as much as the
non-linear heads, so the raw activation scale carries signal.

\begin{table}[!ht]
\centering\small
\caption{Head form on the deployed features and calibration rows.
OOF is the five-fold out-of-fold AUC on the 5{,}228 calibration rows.
Internal and the English mean are AUCs on a separate native feature-dump
run of the evaluation queries.}
\label{tab:headform}
\begin{tabular}{lrrr}
\toprule
Head & OOF & Internal & English mean \\
\midrule
Logistic, raw features (deployed) & 0.848 & 0.876 & 0.771 \\
Logistic, standardized & 0.818 & 0.838 & 0.704 \\
Elastic net, standardized & 0.843 & 0.884 & 0.764 \\
MLP, 256 ReLU units, standardized & 0.815 & 0.819 & 0.702 \\
PCA-256 and gradient-boosted trees & 0.825 & 0.845 & 0.715 \\
\bottomrule
\end{tabular}
\end{table}

\subsection{Deployment ablation: readout versus text router}
\label{app:llmdeploy}
Here we swap only the escalation decision, from the readout to the
zero-shot LLM router, and keep the aggressive budget, the expert, and the relay
fixed. The readout decides earlier and needs no network call to
decide. How much earlier depends on how the router is
scheduled, so we replay two schedules on the main run and then run the
better one live (Figure~\ref{fig:dualview}, M).

\paragraph{Latency, replayed.} The readout decides at the commit to
speak (p50 1.23\,s after the last input sample in the main aggressive
run, p90 1.98\,s) from hidden states the model has already computed
(Figure~\ref{fig:llmtimeline}a). A text router needs a transcript and
an LLM call. We give it the same ASR start rule as the gate arms of
the main run: ASR on the full utterance starts at the scheduled input
end (p50 0.66\,s, Appendix~\ref{app:ttfa-decomp}). This rule uses the
scheduled input end, which a live system does not know, and it does so
for both arms. Thus the comparison measures server TTFA under controlled
replay scheduling, not the full wait after endpoint detection in a live
conversation. Two schedules remain for the call. If the call waits
for the commit, the router decides at $\max(\text{commit}, \text{input
end} + \text{ASR}) + \text{LLM}$. The
router reads only the transcript, so the call can also start as soon
as the transcript is ready, with the decision waiting for whichever of
the commit and the call ends last: $\max(\text{commit}, \text{input
end} + \text{ASR} + \text{LLM})$ (Figure~\ref{fig:llmtimeline}b). The
second schedule removes $\min(\text{LLM}, (\text{commit} - \text{input
end} - \text{ASR})_+)$ of waiting per session. We replay the
per-session timestamps of the main aggressive run (1{,}181 completed
sessions) with the decision moved to each point. Locally answered
turns hold their first audio until the decision, and escalated turns start
the expert later by the same delay. We assume that the router escalates as many sessions as the readout, and the live run below measures the router directly. With the call after the commit,
mean TTFA rises from 4.44\,s to 5.34\,s at the LLM's sequential
latency from our serving region (p50 0.70\,s, p90 0.95\,s over 40 warm
calls) and to 5.41\,s at its latency under eight concurrent calls
(Table~\ref{tab:llmdeploy}). The delay is paid on every turn,
including the 51\% of turns answered locally, which with the readout
start speaking at the commit. With the call at the transcript, mean
TTFA rises to 4.96\,s and 5.00\,s, and only 54\% of turns wait at all,
the rest having both the transcript and the call before the commit.

\begin{table}[!ht]
\centering\small
\caption{Mean TTFA when the escalation decision is made by the readout
or by the zero-shot \texttt{gpt-5.4-mini} router on the ASR transcript,
replayed on the main
aggressive run (1{,}181 completed sessions, five pools, non-negative
TTFA). This is a pooled session mean, not the per-pool macro average
of the main table. The router's ASR starts at the scheduled input end
like the gate arm's. The LLM call either waits for the commit or starts as soon as
the transcript is ready, with the decision waiting for both. LLM
latencies are measured medians, except the assumed 1\,s row.}
\label{tab:llmdeploy}
\begin{tabular}{lrrr}
\toprule
Escalation decision & Call p50 (s) & \multicolumn{2}{c}{Mean TTFA (s), $\Delta$ in parentheses} \\
 & & after the commit & at the transcript \\
\midrule
Hidden-state readout (deployed) & none & \multicolumn{2}{c}{4.44} \\
GPT-5.4 mini, serving region, sequential & 0.70 & 5.34 ($+0.90$) & 4.96 ($+0.52$) \\
GPT-5.4 mini, serving region, 8 concurrent & 0.77 & 5.41 ($+0.97$) & 5.00 ($+0.56$) \\
GPT-5.4 mini, degraded network (assumed) & 1.00 & 5.64 ($+1.20$) & 5.14 ($+0.70$) \\
\bottomrule
\end{tabular}

\end{table}

\paragraph{Live comparison.} We then ran the router live as the
escalation decision on Internal, Speech TriviaQA and SD-QA at the
aggressive budget. The router arm uses zero-shot
GPT-5.4 mini as it is. Its small\_fails probability is thresholded like
the gate's, at the 50\% quantile of the English calibration rows, and
it gets the schedule of Figure~\ref{fig:llmtimeline}b: ASR on the
full utterance from the scheduled input end, exactly as on the gate
arms, the call as soon as the transcript is ready, and the decision at
the later of the commit and the call. On escalation the expert starts
at that moment. Table~\ref{tab:minirouter} pairs the two arms by
query. Inside the pipeline the call took p50 0.76--0.85\,s, and the
router had decided by the commit on only 14--28\% of turns. On the
other turns its decision came p50 0.67--1.29\,s after the commit. The
zero-shot router arm has 77.3\% macro accuracy against the gate's 80.8\%
($-$3.5 points). Its headline latency cost is the extra
wait on the 202 turns that both arms answer locally: 0.64\,s
(paired-bootstrap 95\% interval [0.53, 0.75]), with extra waiting on
every pool. This subset fixes the routing outcome to local answering;
it is not an estimate of the overall mean latency benefit.
Over all 687 completed query pairs, mean TTFA is 4.447\,s for the gate
and 4.457\,s for the router: a router-minus-gate difference of
$+$0.01\,s (95\% interval [$-$0.82, $+$0.59]). The paired median
difference is $+$0.62\,s. We therefore find lower waiting on the
local-answer path, but no clear reduction in the pooled all-pair mean
relative to the text router.

\paragraph{Long waits and sensitivity analysis.}
The full means above include two long
gate-arm sessions on Speech TriviaQA. Their logged delays arise at
different stages. For \texttt{striviaqa0157}, the local model committed
232.65\,s after input end; the subsequent expert request took only
1.78\,s, giving 235.27\,s TTFA. For \texttt{striviaqa0174}, the commit
took 0.87\,s, but the GPT-5.5 request took 102.42\,s, giving
104.18\,s TTFA. The first delay precedes the expert request; the logs
do not establish whether the second reflects API or network variability,
reasoning, or search. Neither session is discarded from the main
results. As a sensitivity analysis, excluding a query pair if either
arm's TTFA exceeds 60\,s leaves 685 pairs and gives a mean difference
of $+$0.46\,s (95\% interval [$+$0.20, $+$0.71]). This removes whole
pairs, rather than capping waits at 60\,s, and shows the influence of
the two long waits. The 202-pair local-answer comparison requires no
60\,s exclusion.

\paragraph{Figure cohorts and denominators.}
Figure~\ref{fig:dualview}'s top row uses the main run
(\texttt{ttfa2}). Its bottom row uses separate deployment runs:
\texttt{ttfa2tr1c} for L/W, \texttt{ttfa2tr1n} for C/B/A, and
\texttt{ttfa2tr2z} for M. These are separate recorded sessions on the
same query IDs, not simultaneous measurements; M was collected about
13 hours after its A control. Each point uses accuracy and timing from
its own run. Accuracy retains all attempted queries, counting failed
sessions as incorrect; timing uses completed sessions with answer
timestamps (Table~\ref{tab:deployment-cohorts}). The top and bottom W
accuracies are 191/240 versus 189/240 on Internal (79.6\% versus
78.8\%) and 180/200 versus 173/200 on SD-QA (90.0\% versus 86.5\%).
These differences reflect separately generated and judged answers,
not latency filtering. The paired comparison in
Table~\ref{tab:minirouter} retains only queries completed in both A and
M (237/250/200 by pool), so its Internal accuracy denominator differs
from the figure's 240 attempts.

\begin{table}[!ht]
\centering\small
\caption{Bottom-row deployment cohorts in Figure~\ref{fig:dualview}.
Accuracy uses all attempts in each arm; TTFA counts every completed
session with an answer timestamp, except the two long waits.}
\label{tab:deployment-cohorts}
\begin{tabular}{lrrrrrrr}
\toprule
& Accuracy $n$/arm & \multicolumn{6}{c}{TTFA $n$ by arm} \\
Pool & & L & C & B & A & W & M \\
\midrule
Internal & 240 & 239 & 240 & 238 & 238 & 239 & 237 \\
Speech TriviaQA & 250 & 250 & 250 & 250 & 248 & 250 & 250 \\
SD-QA & 200 & 200 & 200 & 200 & 200 & 200 & 200 \\
\bottomrule
\end{tabular}
\end{table}

\begin{table}[!ht]
\centering\small
\caption{Live comparison at the aggressive budget: the gate
versus zero-shot GPT-5.4 mini (small\_fails $\geq$ 0.12, ASR from the
input end, call at the transcript) as the escalation decision, paired
by query. Accuracy is the main judge on
the delivered answer, and TTFA is the non-negative wait after the end
of the input. Deltas are router minus gate. The latency column is the
extra wait on turns neither arm escalated, with their count in parentheses.
The overall row reports macro accuracy across the three pools and
pooled counts and latency. Both arms read the routed session.}
\label{tab:minirouter}
\begin{tabular}{lrrrrrr}
\toprule
Pool & $n$ & \multicolumn{2}{c}{Call rate (\%)} & \multicolumn{2}{c}{Accuracy (\%)} & Extra wait (s), both local \\
 & & gate & router & gate & router ($\Delta$) & router $-$ gate \\
\midrule
Internal & 237 & 51.9 & 54.4 & 71.7 & 68.4 ($-$3.4) & $+$0.79 ($n{=}78$) \\
Speech TriviaQA & 250 & 56.0 & 46.8 & 89.2 & 84.0 ($-$5.2) & $+$0.55 ($n{=}84$) \\
SD-QA & 200 & 65.5 & 63.0 & 81.5 & 79.5 ($-$2.0) & $+$0.56 ($n{=}40$) \\
\midrule
Overall & 687 & & & 80.8 & 77.3 ($-$3.5) & $+$0.64 ($n{=}202$) \\
\bottomrule
\end{tabular}

\end{table}

\paragraph{Matched call rate, same sessions.} In this replay we rank the
sessions of the main run by each score and escalate the top $k$, with every
session's local and GPT-5.5 answers fixed to those the main run
delivered, so every replay starts at L and ends at W.
Figure~\ref{fig:dualview} draws zero-shot GPT-5.4 mini at the gate's C,
B and A call rates (dashed orange). The gate's score is the readout
logged at the commit of the same local-arm session whose answer it
routes, and the router's score is its zero-shot small\_fails
probability on that query's live ASR transcript. The replayed gate
curve lies within 1.3 points of the gate's as-run A points on every
pool. Over the five pools the gate is more accurate by $+$2.6
points at the aggressive rate, the routing-gain difference of
Table~\ref{tab:textrouter}, and by $+$1.9 points averaged over the
three tiers.

\paragraph{Network dependence.} The LLM is a remote service, so its
latency depends on where the call is made and on the provider's load.
Table~\ref{tab:llmdeploy} gives mean TTFA at the latencies we measured
and at an assumed 1\,s call. The readout makes no network call to
decide.

\paragraph{Per-turn cost.} The readout adds one dot product over a
hidden state the model has already computed on device. In the benchmark
pipeline both arms send every turn to cloud ASR at input end, escalated
or not, so that an escalation never waits for the transcript. On top of
that the text route adds an LLM request on every turn, including the
turns the local model answers. The LLM call itself is cheap. At the
published prices (\$0.75 per million input tokens, \$4.50 per million
output tokens) it cost about \$0.30 per 1{,}000 turns in our runs (255
input and 24 output tokens per call), far below the expert (\$0.018
per escalation, Appendix~\ref{app:expertcost}). The cost that matters
is one more external dependency on every turn. The gate can also defer
ASR until it escalates, which keeps locally answered turns on the
device and adds the transcription time to escalated turns. A text
router cannot, because it needs the transcript to decide.

\section{Judging and reproducibility}
\label{app:prompts}
\label{app:repro}
The adequacy judge receives the question, optional reference, and candidate
answer without the system identity. Its verbatim system prompt is:

\begin{quote}\small\itshape
You are a strict, fair grader of a small language model's answers. You
are given a QUESTION, an optional REFERENCE answer, and a CANDIDATE
answer produced by a small model. Decide whether the CANDIDATE is
ADEQUATE, meaning ALL of: (1) factually correct; (2) any reasoning
shown is valid; (3) it actually and completely answers the question
that was asked. For multiple-choice questions the candidate must land
on the correct option (by letter or by unambiguously stating the
option's content). A partially correct, hedged, refused, off-topic, or
truncated answer is NOT adequate. If a REFERENCE is provided, grade
against it. Be strict but do not penalize harmless differences in
format or extra correct detail.
\end{quote}

\noindent
The user turn presents \texttt{QUESTION}, the optional
\texttt{REFERENCE answer}, and \texttt{CANDIDATE answer (small
model)} in that order, ending with ``Grade the candidate.'' The
verdict is returned as strict-schema JSON with fields \texttt{reason}
and \texttt{adequate} at low reasoning effort. The expert's
verbatim system prompt is:

\begin{quote}\small\itshape
You are an expert assistant. Answer the user's question correctly,
directly, and concisely. For a multiple-choice question, state the
correct option letter and its content. (Context: this is a single-turn
escalation from an academic research prototype of a small-to-large
model routing gate; your answer is relayed to the user and used for
system evaluation, not for model training.)
\end{quote}

\noindent
OpenAudioBench pools additionally use the official OpenAudioBench
judge prompt and model verbatim
(\texttt{gpt-4o-2024-08-06}, JSON \texttt{correct/incorrect},
any-alias matching), copied unmodified from the benchmark release and
pinned in the code release. Representation-study
sampling and splits use seed 42. Native benchmark arms use sampled decoding and
one run per query. MiniCPM-o~4.5 runs with PyTorch 2.8.0 and Transformers 4.51.0
on single-H100 instances. The release includes calibration manifests,
out-of-fold scores, thresholds, overlap exclusions, benchmark hashes, and the
scripts needed to rebuild every reported table and figure from the recorded
evaluation data.

\end{document}